\documentclass[10pt, a4paper]{article}
\usepackage{lrec2016}
\usepackage{multirow}
\usepackage{caption}
\usepackage{graphicx}
\usepackage{epstopdf}
\usepackage[latin1]{inputenc}

\title{Named Entity Recognition on Twitter for Turkish using Semi-supervised Learning with Word Embeddings}

\name{Eda Okur\textsuperscript{*$\dagger$}, Hakan Demir\textsuperscript{$\dagger$}, Arzucan \"{O}zg\"{u}r\textsuperscript{$\dagger$}}

\address{\textsuperscript{*}Intel Corporation \\
         Istanbul, Turkey \\
         eda.okur@intel.com\\
         \textsuperscript{$\dagger$}Department of Computer Engineering, Bo\u{g}azi\c{c}i University \\
         Istanbul, Turkey \\
         \{hakan.demir, arzucan.ozgur\}@boun.edu.tr\\}
\abstract{
Recently, due to the increasing popularity of social media, the necessity for extracting information from informal text types, such as microblog texts, has gained significant attention. In this study, we focused on the Named Entity Recognition (NER) problem on informal text types for Turkish. We utilized a semi-supervised learning approach based on neural networks. We applied a fast unsupervised method for learning continuous representations of words in vector space. We made use of these obtained word embeddings, together with language independent features that are engineered to work better on informal text types, for generating a Turkish NER system on microblog texts. We evaluated our Turkish NER system on Twitter messages and achieved better F-score performances than the published results of previously proposed NER systems on Turkish tweets. Since we did not employ any language dependent features, we believe that our method can be easily adapted to microblog texts in other morphologically rich languages. \\ 
\newline 
 }

\begin{document}

\maketitleabstract

\section{Introduction}

Microblogging environments, which allow users to post short messages, have gained increased popularity in the last decade. Twitter, which is one of the most popular microblogging platforms, has become an interesting platform for exchanging ideas, following recent developments and trends, or discussing any possible topic. Since Twitter has an enormously wide range of users with varying interests and sharing preferences, a significant amount of content is being created rapidly. Therefore, mining such platforms can extract valuable information. As a consequence, extracting information from Twitter has become a hot topic of research. For Twitter text mining, one popular research area is opinion mining or sentiment analysis, which is surely useful for companies or political parties to gather information about their services and products \cite{Kokciyan-2013}. Another popular research area is content analysis, or more specifically topic modeling, which is useful for text classification and filtering applications on Twitter \cite{Hong-2010}. Moreover, event monitoring and trend analysis are also other examples of useful application areas on microblog texts \cite{Kireyev-2009}. 

In order to build successful social media analysis applications, it is necessary to employ successful processing tools for Natural Language Processing (NLP) tasks such as Named Entity Recognition (NER). NER is a critical stage for various NLP applications including machine translation, question answering and opinion mining. The aim of NER is to classify and locate atomic elements in a given text into predefined categories like the names of the persons, locations, and organizations (PLOs).

NER on well-written texts is accepted as a solved problem for well-studied languages like English. However, it still needs further work for morphologically rich languages like Turkish due to their complex structure and relatively scarce language processing tools and data sets \cite{Seker-2012}. In addition, most of the NER systems are designed for formal texts. The performance of such systems drops significantly when applied on informal texts. To illustrate, the state-of-the-art Turkish NER system has CoNLL F-score of 91.94\% on news data, but the performance drops to F-score of 19.28\% when this system is adopted to Twitter data \cite{Celikkaya-2013}.
\\
There are several challenges for NER on tweets, which are also summarized in \newcite{Kucuk-2014-1}, due to the very short text length and informal structure of the language used. Missing proper grammar rules and punctuation, lack of capitalization and apostrophes, usage of hashtags, abbreviations, and slang words are some of those challenges. In Twitter, using contracted forms and metonymic expressions instead of full organization or location names is very common as well. The usage of non-diacritic characters and the limited annotated data bring additional challenges for processing Turkish tweets.
\\
Due to the dynamic language used in Twitter, heavy feature engineering is not feasible for Twitter NER. \newcite{Demir-2014} developed a semi-supervised approach for Turkish NER on formal (newswire) text using word embeddings obtained from unlabeled data. They obtained promising results without using any gazetteers and language dependent features. We adopted this approach for informal texts and evaluated it on Turkish tweets, where we achieved the state-of-the-art F-score performance. Our results show that using word embeddings for Twitter NER in Turkish can result in better F-score performance compared to using text normalization as a pre-processing step. In addition, utilizing in-domain word embeddings can be a promising approach for Twitter NER.

\section{Related Work}

There are various important studies of NER on Twitter for English. \newcite{Ritter-2011} presented a two-phase NER system for tweets, T-NER, using Conditional Random Fields (CRF) and including tweet-specific features. \newcite{Liu-2011} proposed a hybrid NER approach based on K-Nearest Neighbors and linear CRF. \newcite{Liu-2012} presented a factor graph-based method for NER on Twitter. \newcite{Li-2012} described an unsupervised approach for tweets, called TwiNER. \newcite{Bontcheva-2013} described an NLP pipeline for tweets, called TwitIE. Very recently, \newcite{Cherry-2015} have shown the effectiveness of Brown clusters and word vectors on Twitter NER for English. 
% * <edaaokur@gmail.com> 2016-02-18T14:58:57.125Z:
%
% > For English NER on tweets
%
% REVIEWER #1: + cherry's twitter ner embedding paper from naacl 2015 really ought have been included
%
% ^ <edaaokur@gmail.com> 2016-02-24T21:12:50.481Z:
%
% EDA: Should we mention this recent paper ( http://www.aclweb.org/anthology/N/N15/N15-1075.pdf ) in the above paragraph, i.e. related work - English NER on tweets?
%
% ^.

For Turkish NER on formal texts, \newcite{Tur-2003} presented the first study with a Hidden Markov Model based approach. \newcite{Tatar-2011} presented an automatic rule learning system. \newcite{Yeniterzi-2011} used CRF for Turkish NER, and \newcite{Kucuk-2012} proposed a hybrid approach. A CRF-based model by \newcite{Seker-2012} is the state-of-the-art Turkish NER system with CoNLL F-score of 91.94\%, using gazetteers. \newcite{Demir-2014} achieved a similar F-score of 91.85\%, without gazetteers and language dependent features, using a semi-supervised model with word embeddings.

For Turkish NER on Twitter, \newcite{Celikkaya-2013} presented the first study by adopting the CRF-based NER of \newcite{Seker-2012} with a text normalizer. \newcite{Kucuk-2014-1} adopted a multilingual rule-based NER by extending the resources for Turkish. \newcite{Kucuk-2014-2} adopted a rule-based approach for Turkish tweets, where diacritics-based expansion to lexical resources and relaxing the capitalization yielded an F-score of 48\% with strict CoNLL-like metric.

\section{NER for Turkish Tweets using Semi-supervised Learning}

To build a NER model with a semi-supervised learning approach on Turkish tweets, we used a neural network based architecture consisting of unsupervised and supervised stages.
% * <edaaokur@gmail.com> 2016-02-18T15:14:59.066Z:
%
% > we used a neural network based architecture
%
% REVIEWER #3: Have you considered using conditional random fields, which is commonly used for NER?
%
% ^ <edaaokur@gmail.com> 2016-02-19T16:32:09.156Z:
%
% EDA: Should we really explain why we used NNs instead of CRFs? If so, how should we do that?
%
% ^.

\subsection{Unsupervised Stage}

In the unsupervised stage, our aim is to learn distributed word representations, or word embeddings, in continuous vector space where semantically similar words are expected to be close to each other. Word vectors trained on large unlabeled Turkish corpus can provide additional knowledge base for NER systems trained with limited amount of labeled data in the supervised stage.
% * <edaaokur@gmail.com> 2016-02-18T15:11:23.384Z:
%
% > distributed word representations, or word embeddings, in continuous vector space where semantically similar words are expected to be close to each other.
%
% REVIEWER #3: Briefly describe the intuition behind word embeddings and the methods for obtaining them to the uninitiated reader.
%
% ^.

A word representation is usually a vector associated with each word, where each dimension represents a feature. The value of each dimension is defined to be representing the amount of activity for that specific feature. A distributed representation represents each word as a dense vector of continuous values. By having lower dimensional dense vectors, and by having real values at each dimension, distributed word representations are helpful to solve the sparsity problem. Distributed word representations are trained with a huge unlabeled corpus using unsupervised learning. If this unlabeled corpus is large enough, then we expect that the distributed word representations will capture the syntactic and semantic properties of each word and this will provide a mechanism to obtain similar representations for semantically and syntactically close words.
% * <edaaokur@gmail.com> 2016-02-19T17:41:29.481Z:
%
% > A word representation is usually a vector associated with each word, where each dimension of this vector is actually representing a feature. The value of each dimension is defined to be representing the amount of activity for that specific feature. Distributed representation is representing each word as a dense vector of continuous values. By having lower dimensional dense vectors, and by having real values at each dimension, distributed word representations are helpful to solve the sparsity problem. On the other hand, distributed word representations are trained with huge amount of unlabeled corpus using unsupervised learning. If this unlabeled corpus is huge enough, then we expect that distributed word representations will capture the syntactic and semantic properties of each word and this will provide a mechanism to obtain similar representations for semantically and syntactically close words.
%
% EDA: New paragraph added to address REVIEWER #3's comment "Briefly describe the intuition behind word embeddings".
%
% ^.

Vector space distributed representations of words are helpful for learning algorithms to reach better results in many NLP tasks, since they provide a method for grouping similar words together. The idea of using distributed word representations in vector space is applied to statistical language modeling for the first time by using a neural network based approach with a significant success by \newcite{Bengio-2003}. The approach is based on learning a distributed representation of each word, where each dimension of such a word embedding represents a hidden feature of this word and is used to capture the word's semantic and grammatical properties. Later on, \newcite{Collobert-2011} proposed to use distributed word representations together with the supervised neural networks and achieved state-of-the art results in different NLP tasks, including NER for English.
% * <edaaokur@gmail.com> 2016-03-02T22:39:10.859Z:
%
% > Vector space distributed representations of words are helpful for learning algorithms to reach better results in many NLP tasks, since it provides a method for grouping similar words together. The idea of using distributed word representations in vector space is applied to statistical language modeling for the first time by using neural network based approach with a significant success by \newcite{Bengio-2003}. The approach is based on learning a distributed representation of each word, where each dimension of such word embedding is representing a hidden feature of this word and it is used to capture this word's semantic and grammatical properties. Later on, \newcite{Collobert-2011} proposed to use distributed word representations together with the supervised neural network and achieved state-of-the art results in different NLP tasks, including NER for English.
%
% EDA: New paragraph added to address REVIEWER #3's comment "Briefly describe the intuition behind word embeddings". 
%
% ^.

We used the public tool, word2vec\footnote{https://code.google.com/p/word2vec/}, released by \newcite{Mikolov-2013} to obtain the word embeddings. Their neural network approach is similar to the feed-forward neural networks \cite{Bengio-2003,Collobert-2011}. To be more precise, the previous words to the current word are encoded in the input layer and then projected to the projection layer with a shared projection matrix. 
%This projection is actually the concatenation of the feature vectors of the previous words. 
After that, the projection is given to the non-linear hidden layer and then the output is given to softmax in order to receive a probability distribution over all the words in the vocabulary. However, as suggested by \newcite{Mikolov-2013}, removing the non-linear hidden layer and making the projection layer shared by all words is much faster, which allowed us to use a larger unlabeled corpus and obtain better word embeddings.

\begin{figure}[!h]
\begin{center}
\includegraphics[scale=0.5]{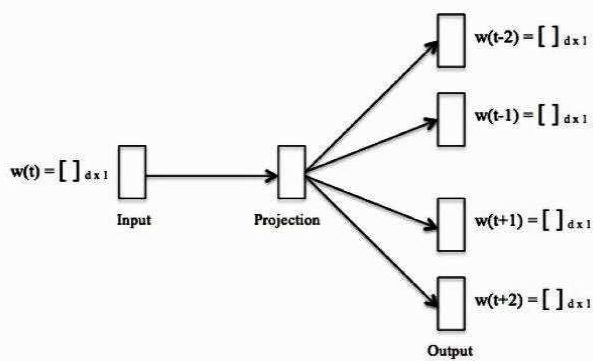}
\caption{Skip-gram model architecture to learn continuous vector representation of words in order to predict surrounding words \cite{Mikolov-2013}.}
\label{fig.1}
\end{center}
\end{figure}

Among the methods presented in \newcite{Mikolov-2013}, we used the continuous Skip-gram model to obtain semantic representations of Turkish words. The Skip-gram model uses the current word as an input to the projection layer with a log-linear classifier and attempts to predict the representation of neighboring words within a certain range. In the Skip-gram model architecture we used, we have chosen 200 as the dimension of the obtained word vectors. The range of surrounding words is chosen to be 5, so that we will predict the distributed representations of the previous 2 words and the next 2 words using the current word. Our vector size and range decisions are aligned with the choices made in the previous study for Turkish NER by \newcite{Demir-2014}. The Skip-gram model architecture we used is shown in Figure \ref{fig.1}. 

\subsection{Supervised Stage}

In this stage, a comparably smaller amount of labeled data is used for training the final NER models. We used the publicly available neural network implementation by \newcite{Turian-2010}\footnote{http://cogcomp.cs.illinois.edu/Data/ACL2010\_NER\_Experim ents.php}, which actually follows the study by \newcite{Ratinov-2009}, where a regularized averaged multiclass perceptron is used. 
% * <edaaokur@gmail.com> 2016-02-17T17:10:41.632Z:
%
% > We used the public neural network implementation by \newcite{Ratinov-2009}\footnote{http://cogcomp.cs.illinois.edu/Data/ACL2010\_NER\_Experim ents.php}
%
% REVIEWER #1: + in section 3.2, the ratinov and roth work is not at all a neural network system. and the tool in the footnote (2) is the turian, ratinov and bengio work wrong citation.
%
% ^.
% * <edaaokur@gmail.com> 2016-03-02T23:41:57.347Z:
%
% > We used the publicly available neural network implementation by \newcite{Turian-2010}\footnote{http://cogcomp.cs.illinois.edu/Data/ACL2010\_NER\_Experim ents.php}, which is actually following a study by \newcite{Ratinov-2009}, where a regularized averaged multiclass perceptron is used.
%
% EDA: This sentence is updated (citation is corrected) to address REVIEWER #1's comment.
%
% ^.

Note that although non-local features are proven to be useful for the NER task on formal text types such as news articles, their usage and benefit is questionable for informal and short text types. Due to the fact that each tweet is treated as a single document with only 140 characters, it is difficult to make use of non-local features such as context aggregation and prediction history for the NER task on tweets. On the other hand, local features are mostly related to the previous and next tokens of the current token. With this motivation, we explored both local and non-local features but observed that we achieve better results without non-local features. As a result, to construct our NER model on tweets, we used the following local features:
% * <edaaokur@gmail.com> 2016-02-18T15:13:44.202Z:
%
% > since non-local features are not suitable for microblog texts
%
% REVIEWER #3: Motivate why "nonlocal features" are not suitable for microblog texts.
%
% ^.
% * <edaaokur@gmail.com> 2016-03-02T23:17:11.563Z:
%
% > Note that although non-local features are proven to be useful for the NER task on formal text types such as news articles, their usage and benefit is questionable for informal and short text types. Due to the fact that each tweet is treated as a single document with only 140 characters, it is difficult to make use of non-local features such as context aggregation and prediction history for the NER task on tweets. With this motivation, we explored both local and non-local features but observed that we achieve better results without non-local features. As a result,
%
% EDA: This paragraph is expanded in order to address REVIEWER #3's comment.
%
% ^.

\begin{itemize}
    \item{Context: All tokens in the current window of size two.}
% * <edaaokur@gmail.com> 2016-02-18T15:19:41.853Z:
%
% > Context: All tokens in the window of size two.
%
% REVIEWER #3: It is unclear how the described features are actually represented, e.g., context as binary features or as a bag of words?
%
% ^ <edaaokur@gmail.com> 2016-03-03T00:14:47.948Z:
%
% EDA: In none of the previous studies using these same NER features, this detail is not mentioned ever. Should we really explain this for each feature? If so, how should we address this comment?
%
% ^.
    \item{Capitalization: Boolean feature indicating whether the first character of a token is upper-case or not. This feature is generated for all the tokens in the current window.}
    \item{Previous tags: Named entity tag predictions of the previous two tokens.}
    \item{Word type information: Type information of tokens in the current window, i.e. all-capitalized, is-capitalized, all-digits, contains-apostrophe, and is-alphanumeric.}
    \item{Token prefixes: First characters with length three and four, if exists, of current token.}
    \item{Token suffixes: Last characters with length one to four, if exists, of current token.}
    \item{Word embeddings: Vector representations of words in the current window.}
\end{itemize}

In addition to tailoring the features used by \newcite{Ratinov-2009} for tweets, there are other Twitter-specific aspects of our NER system such as using word embeddings trained on an unlabeled tweet corpus, applying normalization on labeled tweets, and extracting Twitter-specific keywords like hashtags, mentions, smileys, and URLs from both labeled and unlabeled Turkish tweets. For text normalization as a pre-processing step of our system, we used the Turkish normalization interface\footnote{http://tools.nlp.itu.edu.tr/Normalization} developed for social media text with ill formed word detection and candidate word generation \cite{Torunoglu-2014}.

Along with the features used, the representation scheme for named entities is also important in terms of performance for a NER system. Two popular such encoding schemes are BIO and BILOU. The BIO scheme identifies the Beginning, the Inside and the Outside of the named entities, whereas the BILOU scheme identifies the Beginning, the Inside and the Last tokens of multi-token named entities, plus the Outside if it is not a named entity and the Unit length if the entity has single token. Since it is shown by \newcite{Ratinov-2009} that BILOU representation scheme significantly outperforms the BIO encoding scheme, we make use of BILOU encoding for tagging named entities in our study. Furthermore, we applied normalization to numerical expressions as described in \newcite{Turian-2010}, which helps to achieve a degree of abstraction to numerical expressions.

\section{Data Sets}

\subsection{Unlabeled Data}

In the unsupervised stage, we used two types of unlabeled data to obtain Turkish word embeddings. The first one is a Turkish news-web corpus containing 423M words and 491M tokens, namely the BOUN Web Corpus\footnote{http://79.123.177.209/~hasim/langres/BounWebCorpus.tgz} \cite{Sak-2008,Sak-2011}. The second one is composed of 21M Turkish tweets with 241M words and 293M tokens, where we combined 1M tweets from TS TweetS\footnote{http://tscorpus.com/en} by \newcite{Sezer-2013} and 20M Turkish Tweets\footnote{http://www.kemik.yildiz.edu.tr/data/File/20milyontweet.rar} by Bolat and Amasyal{\i}.

We applied tokenization on both Turkish news-web corpus and Turkish tweets corpus using the publicly available Zemberek\footnote{https://github.com/ahmetaa/zemberek-nlp} tool developed for Turkish. We have also applied lower-casing on both corpora in order to limit the number of unique words. Since our combined tweets corpus is composed of Twitter-specific texts, we applied what we call Twitter processing where we replaced mentions, hashtags, smileys and URLs with certain keywords.
% * <edaaokur@gmail.com> 2016-02-18T15:20:32.211Z:
%
% > by Bolat and Amasyal{\i}.
%
% REVIEWER #3: Is there a (paper) reference to the Twitter dataset released by Bolat and Amasyal?
%
% ^ <edaaokur@gmail.com> 2016-02-24T21:27:55.873Z:
%
% EDA: I've provided the Kemik's 20M Turkish tweets dataset link in the footnote, but I couldn't find a paper reference for this dataset in the website. As written in the readme file provided within this dataset, it is constructed by Anıl BOLAT, M. Fatih AMASYALI. What should we do about this comment? If there is no reference paper provided, should we remove those names and only keep the dataset link in the footnote?
%
% ^.

\subsection{Labeled Data}

In the supervised stage, we used two types of labeled data to train and test our NER models. The first one is Turkish news data annotated with ENAMEX-type named entities, or PLOs \cite{Tur-2003}. It includes 14481 person, 9409 location, and 9034 organization names in the training partition of 450K words. This data set is popularly used for performance evaluation of NER systems for Turkish, including the ones presented by \newcite{Seker-2012}, by \newcite{Yeniterzi-2011} and by \newcite{Demir-2014}. 

The second type of labeled data is annotated Turkish tweets, where we used two different sets. The first set, TwitterDS-1, has around 5K tweets with 54K tokens and 1336 annotated PLOs \cite{Celikkaya-2013}. The second set, TwitterDS-2\footnote{http://optima.jrc.it/Resources/2014\_JRC\_Twitter\_TR\_NER-dataset.zip}, which is publicly available, has 2320 tweets with around 21K tokens and 980 PLOs in total \cite{Kucuk-2014-1}. The counts for each of the ENAMEX-type named entities for these Turkish Twitter data sets are provided in Table \ref{tab.0}.
% * <edaaokur@gmail.com> 2016-02-18T15:24:49.516Z:
%
% > The first set, TwitterDS-1, has 50K tokens and 1336 PLO tokens \cite{Celikkaya-2013}. The second set, TwitterDS-2\footnote{http://optima.jrc.it/Resources/2014\_JRC\_Twitter\_TR\_NER-dataset.zip}, has 21K tokens and 980 PLO phrases \cite{Kucuk-2014-1}.
%
% REVIEWER #3: Why is one dataset described according to the number of tokens (what is the number of NE mentions?) and the other as the number of phrases?
%
% ^.
% * <edaaokur@gmail.com> 2016-02-24T22:01:37.330Z:
%
% > The first set, TwitterDS-1, has around 5K tweets with 54K tokens and 1336 annotated PLOs \cite{Celikkaya-2013}. The second set, TwitterDS-2\footnote{http://optima.jrc.it/Resources/2014\_JRC\_Twitter\_TR\_NER-dataset.zip}, has 2320 tweets with around 21K tokens and 980 PLOs in total \cite{Kucuk-2014-1}.
%
% EDA: The tweets datasets explanation is updated (tokens-phrases part is corrected) & expanded to address REVIEWER #3's comment.
%
% ^.

\begin{table}[!h]
\begin{center}
\begin{tabular}{|l|c|c|}

      \hline
      & \textbf{Twitter DS-1} & \textbf{Twitter DS-2} \\
      & (TwtDS-1) & (TwtDS-2) \\
      \hline\hline
      \textbf{Data Size (\#tokens)} & 54K & 21K \\
      \hline\hline
      Person & 676 & 457 \\
      \hline
      Location & 241 & 282 \\
      \hline
      Organization & 419 & 241 \\
      \hline\hline
      \textbf{Total PLOs} & 1336 & 980 \\
      \hline

\end{tabular}
\captionsetup{justification=centering}
\caption{Number of PLOs in Turkish Twitter data sets.}
\label{tab.0}
\end{center}
\end{table}

\section{Experiments and Results}
% * <edaaokur@gmail.com> 2016-02-18T15:35:08.902Z:
%
% > Experiments and Results
%
% REVIEWER #3: Significance testing would strengthen the paper.
%
% ^ <edaaokur@gmail.com> 2016-02-24T22:08:16.574Z:
%
% EDA: What should we do about this comment?
%
% ^.

We designed a number of experimental settings to investigate their effects on Turkish Twitter NER. These settings are as follows: the text type of annotated data used for training, the text type of unlabeled data used to learn the word embeddings, using the capitalization feature or not, and applying text normalization. We evaluated all models on ENAMEX types with the CoNLL metric and reported phrase-level overall F-score performance results. To be more precise, the F-score values presented in Table \ref{tab.1}, Table \ref{tab.2} and Table \ref{tab.3} are micro-averaged over the classes using the strict metric.
% * <edaaokur@gmail.com> 2016-02-18T14:44:11.408Z:
%
% > We evaluated all models on ENAMEX types with the CoNLL metric and reported phrase-level overall F-score performances.
%
% REVIEWER #1: + what metric is used in table 1, table 2? strict, lenient or average chunk f1?
%
% REVIEWER #3: Are the F1?scores macro- or micro-averaged over the classes?
%
%
% ^.
% * <edaaokur@gmail.com> 2016-02-24T22:20:42.231Z:
%
% > As a result, F-score values presented in Table 1 and Table 2 are micro-averaged over the classes in strict metric.
%
% EDA: I think the previous sentence already answers REVIEWER #1's and REVIEWER #3's comments, since CoNLL uses micro-averaging and it is a strict metric (phrase-level). In any case, I've added this last sentence to further clarify the metrics we used in results tables, in order to address REVIEWER #1's and REVIEWER #3's comments. 
%
% ^.

\subsection{NER Models Trained on News}

Most of our NER models are trained on annotated Turkish news data by \newcite{Tur-2003} and tested on tweets, due to the limited amount of annotated Turkish tweets.
% * <edaaokur@gmail.com> 2016-02-18T14:42:56.888Z:
%
% > Most of our NER models are trained on annotated Turkish news data by \newcite{Tur-2003} and tested on tweets, due to the limited amount of annotated Turkish tweets.
%
% REVIEWER #1: + training on news and evaluating over social media is weird this makes the results lowvalue and only able to make general claims at low performance levels. the prime question is what effect indomain data has, and this goes unanswered, because no new data was presented with this paper.
%
% ^ <edaaokur@gmail.com> 2016-03-03T00:16:49.075Z:
%
% EDA: I think we have nothing to do now about this comment...
%
% ^.

\begin{table}[!h]
\begin{center}
\begin{tabular}{|l||c||c|c|c|}

\hline
\multirow{2}{*}{\textbf{Test Set}} & \multirow{2}{*}{\textbf{Cap}} & 
\multicolumn{3}{c|}{\textbf{Phrase-level (Overall)}} \\ 
\cline{3-5} 
 									&	  & \textbf{Web} & \textbf{Twt} & \textbf{W+T} \\ 
\hline\hline
\multirow{2}{*}{TwtDS-1} 	   		& ON  & 36.55 & 35.14 & 38.11 \\ 
\cline{2-5} 
                   			       	& OFF & 38.52 & 31.57 & \textbf{40.18} \\ 
\hline
\multirow{2}{*}{TwtDS-1\_Norm} 		& ON  & 41.82 & 41.54 & \underline{\textbf{42.79}} \\ \cline{2-5} 
                   				   	& OFF & 40.50 & 39.31 & 41.04 \\ 
\hline\hline
\multirow{2}{*}{TwtDS-1\_FT} 	   	& ON  & 40.53 & 40.44 & 41.86 \\ 
\cline{2-5} 
                   			        & OFF & 41.63 & 36.43 & \textbf{44.00} \\ 
\hline
\multirow{2}{*}{TwtDS-1\_FT\_Norm} 	& ON  & 45.74 & 46.27 & \underline{\textbf{46.61}} \\ \cline{2-5} 
                   				    & OFF & 44.17 & 44.91 & 45.27 \\ 
\hline\hline
\multirow{2}{*}{TwtDS-2} 	   		& ON  & 53.14 & 47.72 & 54.01 \\ 
\cline{2-5} 
                   			        & OFF & 54.09 & 48.15 & \textbf{55.45} \\ 
\hline
\multirow{2}{*}{TwtDS-2\_Norm} 		& ON  & 55.20 & 52.23 & \underline{\textbf{56.79}} \\ 
\cline{2-5} 
                   				   	& OFF & 54.75 & 49.43 & 56.12 \\ 
\hline
                                        
  % 36.55	35.14	38.11
  % 38.52	31.57	40.18
  % 41.82	41.54	42.79
  % 40.50	39.31	41.04
  % 40.53	40.44	41.86
  % 41.63	36.43	44.00
  % 45.74	46.27	46.61
  % 44.17	44.91	45.27
  % 53.14	47.72	54.01
  % 54.09	48.15	55.45
  % 55.20	52.23	56.79
  % 54.75	49.43	56.12

\end{tabular}
\captionsetup{justification=centering}
\caption{Phrase-level overall F-score performance results of the NER models trained on news.}
\label{tab.1}
\end{center}
\end{table}

In addition to using TwitterDS-1 and TwitterDS-2 as test sets, we detected 291 completely non-Turkish tweets out of 5040 in TwitterDS-1 and filtered them out using the isTurkish\footnote{http://tools.nlp.itu.edu.tr/IsTurkish} tool \cite{Sahin-2013} to obtain TwitterDS-1\_FT. We also used the normalized versions of these data sets. As shown in Table \ref{tab.1}, turning off the capitalization feature is better when text normalization is not applied (bold entries), but the best results are achieved when normalization is applied and the capitalization feature is used (underlined bold entries). To observe the effects of the type of the source text used to learn the word embeddings, we have three models as Web, Twt, and Web+Twt where we used the Turkish web corpus, tweet corpus, and their combination respectively to learn the word embeddings. Including in-domain data from a relatively smaller tweet corpus together with a larger web corpus yields in better Twitter NER performance.
% * <edaaokur@gmail.com> 2016-02-18T14:47:53.120Z:
%
% > Including in-domain data from a relatively smaller tweet corpus together with a larger web corpus yields in better Twitter NER performance.
%
% REVIEWER #1: + when generating the embeddings, although genre has been changed, so has the corpus size. without control for dataset size, there are two variables moving at the same time, and so this doesn't tell us enough. we already know that bigger datasets build better unsupervised representations for NER (e.g. the Brown clustering tuning paper from RANLP this year). so all you can REALLY conclude is that "either genrematching or bigger datasets, help" which is uninteresting we already know that more data = better!
%
% ^ <edaaokur@gmail.com> 2016-03-03T00:17:23.889Z:
%
% EDA: I think we have nothing to do now about this comment...
%
% ^.

\subsubsection{Word Embeddings versus Text Normalization}

We examined the effects of word embeddings on the performance of our NER models, and compared them to the improvements achieved by applying normalization on Turkish tweets. The baseline NER model is built by using the features explained in section 3.2, except the capitalization and word embeddings features. Using word embeddings obtained with unsupervised learning from a large corpus of web articles and tweets results in better NER performance than applying a Twitter-specific text normalizer, as shown in Table \ref{tab.2}. This is crucial since Turkish text normalization for unstructured data is a challenging task and requires successful morphological analysis, whereas extracting word embeddings for any language or domain is much easier, yet more effective.
% * <edaaokur@gmail.com> 2016-02-18T14:54:02.538Z:
%
% > Using word embeddings obtained with unsupervised learning from a large corpus of web articles and tweets results in better NER performance than applying a Twitter-specific text normalizer, as shown in Table 2.
%
% REVIEWER #1: + there's a big jump in table 2 in the +cap data, on DS1_ FT: compare BL+Cap+Norm , BL+Cap+WordE , BL+Cap+WordE+Norm. What's going on there? Can you tell us? It looks like Norm is as useful as WordE which might be no surprise, given the small datasets used, the variation in tweets leader to sparser vocabularies, and the resulting sparsity in the induced clusterings. This really is the smoking gun that not enough data has been used in the paper.
%
% ^ <edaaokur@gmail.com> 2016-03-03T00:19:45.241Z:
%
% EDA: I think we have nothing to do now about this comment, since we cannot increase the data size we used. Should we still explain further the results we see in table 2, to address this comment's "What's going on there? Can you tell us?" part?
%
% ^.

\begin{table}[!h]
\begin{center}
\begin{tabular}{|l||c|c|c|}

\hline
\multirow{3}{*}{\textbf{NER Model}} & 
\multicolumn{3}{c|}{\textbf{Phrase-level (Overall)}} \\ 
\cline{2-4} 
 & \textbf{Twt} & \textbf{Twt} & \textbf{Twt} \\ 
 & \textbf{DS-1} & \textbf{DS-1\_FT} & \textbf{DS-2} \\ 
\hline\hline
Baseline(BL) 				& 22.16 & 25.98 & 35.16 \\ \hline 
BL+\textbf{Norm} 			& 33.05 & 39.23 & 37.17 \\ \hline
BL+\textbf{WordE} 			& 40.18 & 44.00 & 55.45 \\ \hline
BL+\textbf{WordE+Norm} 		& \textbf{41.04} & \textbf{45.27} & \textbf{56.12} \\ \hline\hline
Baseline(BL)+Cap 			& 27.16 & 30.21 & 37.32 \\ \hline 
BL+Cap+\textbf{Norm} 		& 36.70 & 40.78 & 42.18 \\ \hline
BL+Cap+\textbf{WordE} 		& 38.11 & 41.86 & 54.01 \\ \hline
BL+Cap+\textbf{WordE+Norm} 	& \textbf{42.79} & \textbf{46.61} & \textbf{56.79} \\ \hline
                                        
  % 22.16	25.98	35.16
  % 33.05	39.23	37.17
  % 40.18	44.00	55.45
  % 41.04	45.27	56.12
  % 27.16	30.21	37.32
  % 36.70	40.78	42.18
  % 38.11	41.86	54.01
  % 42.79	46.61	56.79

\end{tabular}
\captionsetup{justification=centering}
\caption{Phrase-level overall F-score performance results to compare word embeddings and normalization.}
\label{tab.2}
\end{center}
\end{table}

\begin{table*}[ht]
\begin{center}
\begin{tabular}{|c||c||c|c|c|c||c||c|}

\hline
\multirow{2}{*}{\textbf{System}} & 
\multirow{2}{*}{\textbf{Trained On}} & 
\multicolumn{4}{c||}{\textbf{Best Settings}} &
\multirow{2}{*}{\textbf{Test Set}} &
\textbf{Phrase-level} \\ 
\cline{3-6} 
 & & \textbf{Gazet} & \textbf{Norm} & \textbf{Cap} & \textbf{Other} &
& \textbf{(Overall)} \\
\hline\hline
\c{C}elikkaya et al., & Turkish News  & \multirow{2}{*}{Yes} & \multirow{2}{*}{Yes} & \multirow{2}{*}{ON} & \multirow{2}{*}{-} & \multirow{2}{*}{TwtDS-1} & \multirow{2}{*}{19.28} \\
(2013) & (T\"{u}r et al., 2003) & & & & & & \\
\hline
K\"{u}\c{c}\"{u}k et al., & \multirow{2}{*}{EMM News} & \multirow{2}{*}{Yes} & \multirow{2}{*}{No} & \multirow{2}{*}{ON} & relaxed \& extended & TwtDS-1 & 36.11 \\
\cline{7-8} 
(2014) & & & & & gazetteer & TwtDS-2 & 42.68 \\
\hline
K\"{u}\c{c}\"{u}k and & \multirow{2}{*}{no training} & \multirow{2}{*}{Yes} & \multirow{2}{*}{No} & \multirow{2}{*}{OFF} & diacritics expanded & \textit{TwtDS-1} & \textit{38.01} \\
\cline{7-8} 
Steinberger, (2014) & & & & & gazetteer & \textit{TwtDS-2} & \textit{48.13} \\
\hline
\multirow{5}{*}{Our NER Systems} & \multirow{3}{*}{\begin{tabular}{@{}c@{}}Turkish News\\(T\"{u}r et al., 2003)\end{tabular}} & \multirow{3}{*}{No} & \multirow{3}{*}{Yes} & \multirow{3}{*}{ON} & word embeddings + & \multirow{2}{*}{TwtDS-1} & \multirow{2}{*}{46.61} \\
 & & & & & filter non-Turkish & & \\
\cline{6-8} 
 & & & & & word embeddings & \textbf{TwtDS-2} & \textbf{56.79} \\
\cline{2-8}
 & Turkish Tweets & \multirow{2}{*}{No} & \multirow{2}{*}{Yes} & \multirow{2}{*}{ON} & word embeddings + & \multirow{2}{*}{\textbf{TwtDS-1}} & \multirow{2}{*}{\textbf{48.96}} \\
 & (TwtDS-2) & & & & filter non-Turkish & & \\
\hline

\end{tabular}
\caption{Phrase-level overall F-score performance results compared to the state-of-the-art.}
\label{tab.3}
\end{center}
\end{table*}

\subsection{NER Models Trained on Tweets}

Although an ideal Turkish NER model for Twitter should be trained on similar informal texts, all previous Turkish Twitter NER systems are trained on news data due to the limited amount of annotated Turkish tweets. We also experimented training NER models on relatively smaller labeled Twitter data with 10-fold cross-validation. Our best phrase-level F-score of 46.61\% achieved on TwitterDS-1\_FT is increased to 48.96\% when trained on the much smaller tweets data, TwitterDS-2, instead of news data.

\subsection{Comparison with the State-of-the-art}
% * <edaaokur@gmail.com> 2016-02-17T16:59:49.268Z:
%
% > Comparison with the State-of-the-art
%
% REVIEWER #2: there is some lacks of informations in the state of the art: authors should discuss other approachs instead of enumerate them and link them with their own result more explicitly. With this aim, Table 3 is very clear and helpful in the results section.
%
% ^.

The best F-scores of the previously published Turkish Twitter NER systems \cite{Celikkaya-2013,Kucuk-2014-1,Kucuk-2014-2} as well as our proposed NER system are shown in Table \ref{tab.3}. We used the same training set with the first system \cite{Celikkaya-2013} in our study, but the second NER system \cite{Kucuk-2014-1} uses a different multilingual news data and the third system \cite{Kucuk-2014-2}, which is rule based, does not have a training phase at all. All of these previous NER systems use gazetteer lists for named entities, which are manually constructed and highly language-dependent, whereas our system does not. Note that there is no publicly available gazetteer lists in Turkish. \newcite{Kucuk-2014-2} achieved the state-of-the-art performance results for Turkish Twitter NER with their best model settings (shown in italic). These settings are namely using gazetteers list, with capitalization feature turned off, and with no normalization, together by expanding their gazetteer lists of named entities with diacritics variations. 

Our proposed system outperforms the state-of-the-art results on both Turkish Twitter data sets, even without using gazetteers (shown in bold). We achieved our best performance results with Turkish word embeddings obtained from our Web+Tweets corpus, when we apply normalization on tweets and keep the capitalization as a feature.
% * <edaaokur@gmail.com> 2016-03-03T01:14:02.340Z:
%
% > We used the same training set with the first system \cite{Celikkaya-2013} in our study, but the second NER system \cite{Kucuk-2014-1} uses a different multilingual news data and the third system \cite{Kucuk-2014-2}, which is rule based, does not have a training phase at all. All of these previous NER systems use gazetteer lists for named entities, which are manually constructed and highly language-dependent, whereas our system does not. Note that there is no publicly available gazetteer lists in Turkish. \newcite{Kucuk-2014-2} achieved the state-of-the-art performance results for Turkish Twitter NER with their best model settings. These settings are namely using gazetteers list, with capitalization feature turned off, and with no normalization, together by expanding their gazetteer lists of named entities with diacritics variations. Our proposed system outperforms the state-of-the-art results on both data sets, even without using gazetteers. We achieved our best performance results with Turkish word embeddings obtained from our Web+Tweets corpus, when we apply normalization on tweets and keep the capitalization as a feature.
%
% EDA: This paragraph is expanded to address REVIEWER #2's comment.
%
% ^.

\section{Conclusion}

We adopted a neural networks based semi-supervised approach using word embeddings for the NER task on Turkish tweets. At the first stage, we attained distributed representations of words by employing a fast unsupervised learning method on a large unlabeled corpus. At the second stage, we exploited these word embeddings together with language independent features in order to train our neural network on labeled data. We compared our results on two different Turkish Twitter data sets with the state-of-the-art NER systems proposed for Twitter data in Turkish and showed that our system outperforms the state-of-the-art results on both data sets. Our results also show that using word embeddings from an unlabeled corpus can lead to better performance than applying Twitter-specific text normalization. We discussed the promising benefits of using in-domain data to learn word embeddings at the unsupervised stage as well. Since the only language dependent part of our Turkish Twitter NER system is text normalization, and since even without text normalization it outperforms the previous state-of-the-art results, we believe that our approach can be adapted to other morphologically rich languages. Our Turkish Twitter NER system, namely TTNER, is publicly available\footnote{http://tabilab.cmpe.boun.edu.tr/projects/ttner/}.

We believe that there is still room for improvement for NLP tasks on Turkish social media data. As a future work, we aim to construct a much larger in-domain resource, i.e., unlabeled Turkish tweets corpus, and investigate the full benefits of attaining word embeddings from in-domain data on Twitter NER.

% * <edaaokur@gmail.com> 2016-02-18T15:01:16.534Z:
%
% REVIEWER #3: The ability to leverage outdomain resources for learning word embeddings is interesting; however, in future work it be interesting to investigate whether using a large(r) indomain resource (i.e., an unannotated Twitter corpus) would produce better results.
%
% ^.

\section{Acknowledgements}

This research is partially supported by Bo\u{g}azi\c{c}i University Research Fund Grant Number 11170. We would also like to thank The Scientific and Technological Research Council of Turkey (T\"{U}B\.{I}TAK), The Science Fellowships and Grant Programmes Department (B\.{I}DEB) for providing financial support with 2210 National Scholarship Programme for MSc Students.

% \nocite{*}
\section{Bibliographical References}
\label{main:ref}

\bibliographystyle{lrec2016}
\bibliography{xample}

% \section{Language Resource References}
% \label{lr:ref}
% \bibliographystylelanguageresource{lrec2016}
% \bibliographylanguageresource{xample}

\end{document}